\pdfoutput=1
\newif\ifieeetran
\IfFileExists{IEEEtran.cls}{\ieeetrantrue}{\ieeetranfalse}
\ifieeetran
  \documentclass[conference]{IEEEtran}
  \IEEEoverridecommandlockouts
\else
  \documentclass[letterpaper,10pt,twocolumn]{article}
  \usepackage[left=0.68in,right=0.68in,top=0.72in,bottom=0.82in,columnsep=0.25in]{geometry}
\fi

\usepackage{amsmath,amssymb}
\usepackage{array}
\usepackage{booktabs}
\usepackage{cite}
\usepackage{graphicx}
\usepackage{placeins}
\usepackage{url}

\ifieeetran\else
  \newenvironment{IEEEkeywords}{%
    \par\smallskip\noindent\textbf{Index Terms---}\ignorespaces%
  }{\par\smallskip}
\fi

\graphicspath{{figures/}}

\begin{document}

\title{RAGrasp: Geometry-Semantic Template Retrieval and Grasp Transfer}

\ifieeetran
\author{
    \IEEEauthorblockN{
        Shenzhe Zhu\IEEEauthorrefmark{1}\IEEEauthorrefmark{2},
        Chengxiao He\IEEEauthorrefmark{1}\IEEEauthorrefmark{2}, and
        Jan Harder\IEEEauthorrefmark{1}\IEEEauthorrefmark{3}
    }
    \IEEEauthorblockA{
        \IEEEauthorrefmark{1}Tongji University\\
        Email: \{2531819, 2130238, harderjan\}@tongji.edu.cn
    }
    \IEEEauthorblockA{
        \IEEEauthorrefmark{2}These authors contributed equally.
    }
    \IEEEauthorblockA{
        \IEEEauthorrefmark{3}Corresponding author: Jan Harder
        (harderjan@tongji.edu.cn).
    }
}
\else
\author{
    Shenzhe Zhu\textsuperscript{1,*}\quad
    Chengxiao He\textsuperscript{1,*}\quad
    Jan Harder\textsuperscript{1,\dagger}\\[0.5em]
    \textsuperscript{1}Tongji University\\
    \textsuperscript{*}These authors contributed equally.\\
    \textsuperscript{\dagger}Corresponding author:
    \texttt{harderjan@tongji.edu.cn}
}
\date{}
\fi

\maketitle
\pagestyle{plain}
\thispagestyle{plain}

\begin{abstract}
We present RAGrasp, a retrieval-augmented pipeline for planar parallel-jaw grasping from a compact set of locally collected, grasp-annotated RGB-D (color and depth) templates. Unlike task-specific predictors trained primarily on large public or synthetic grasp datasets, RAGrasp requires no end-to-end retraining for a new deployment.Its template memory is constructed from observations collected with the deployment camera, robot, and gripper in the target workspace, thereby aligning stored examples with the local sensing and embodiment conditions. The system uses self-supervised DINOv2 visual features together with appearance and depth cues to prompt the Segment Anything Model~2 (SAM2), which isolates the query object. A two-stage geometry-semantic retrieval cascade then selects a template, and a confidence gate chooses one of two grasp-transfer estimators. The transferred grasp is refined using mask-support and silhouette-contact constraints before calibrated 2D-to-3D conversion. In real-world trials, RAGrasp achieves 20/20 successful grasps on seen objects and 19/20 on unseen objects. Within the evaluated setting, the results demonstrate deployment-specific grasp adaptation from limited local annotation and tolerance to the tested viewpoint and illumination changes.
\end{abstract}


\section{Introduction}
Robotic grasping systems must remain reliable when object identity, pose, illumination, camera viewpoint, and hardware differ from the conditions represented during training. Although large public and synthetic datasets enable scalable end-to-end grasp learning, their visual statistics, sensor noise, robot embodiment, gripper geometry, and workspace configuration may not match a specific deployment. This domain gap can degrade grasp precision on the target system. Collecting a comparably large local dataset and retraining or fine-tuning a predictor for every installation is costly. The practical problem addressed in this work is therefore to achieve accurate grasping from only a small number of locally acquired and annotated examples.

RAGrasp addresses this problem through retrieval and adaptation, drawing inspiration from retrieval-augmented models that separate external memory from inference~\cite{lewis2020rag}. Rather than learning a task-specific grasp predictor from a public corpus, it builds a compact template bank from RGB-D observations collected with the deployment hardware. Given a new observation, the method isolates the query object, retrieves compatible local templates using geometric and semantic evidence, transfers the stored grasp through a confidence-gated dual-path estimator, and refines the result using mask-support and silhouette-contact constraints. A verified execution can optionally be stored as a new template. This design separates reusable grasp experience from the transfer procedure and expands deployment coverage without retraining the full system.

The main contributions are:
\begin{itemize}
  \item a data-efficient grasping formulation that uses a small, locally collected template bank instead of task-specific training on a large public grasp dataset;
  \item a geometry-semantic retrieval cascade that combines scaled multi-start contour ICP with masked DINOv2 ranking;
  \item a confidence-gated dual transfer mechanism that chooses between contour alignment and feature-correspondence-based similarity estimation;
  \item a mask-constrained local refinement stage followed by calibrated metric execution and success-conditioned memory write-back; and
  \item a real-world evaluation across seen and unseen objects, viewpoint and illumination changes, template-bank sizes, similarity strata, and component ablations.
\end{itemize}

\begin{figure*}[t]
  \centering
  \includegraphics[width=\textwidth]{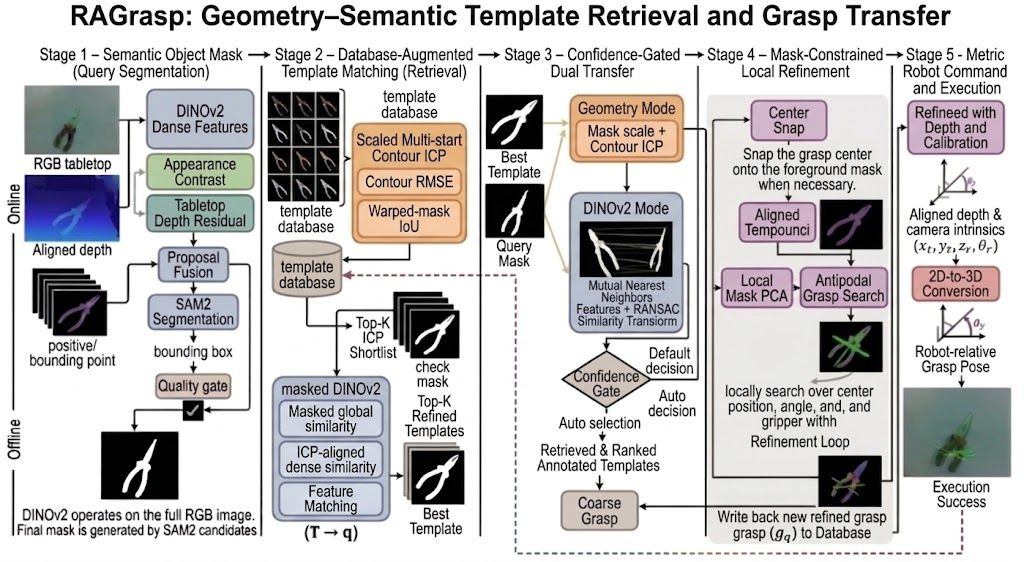}
  \caption{Overview of the RAGrasp pipeline. Stage 1 forms a quality-gated SAM2 query mask from RGB-D evidence. Stage 2 performs contour-ICP shortlisting and DINOv2-based fine ranking. Stage 3 selects a geometric or semantic transfer path using a confidence gate. Stage 4 performs mask-constrained local grasp refinement and optional success-conditioned memory update. Stage 5 converts the refined image-space grasp into a calibrated robot command.}
  \label{fig:main-pipeline}
\end{figure*}

\section{Related Work}
\subsection{Learning-Based Grasp Detection}
Classical grasp analysis studies force closure, grasp quality, and simulation~\cite{nguyen1988forceclosure,ferrari1992optimal,miller2004graspit}; comprehensive surveys trace the transition from analytic geometry to data-driven synthesis~\cite{bohg2014survey,newbury2023review}. Early vision-based learning demonstrated grasp transfer to novel objects~\cite{saxena2008robotic}. More recent methods include analytic-label learning in Dex-Net~2.0~\cite{mahler2017dexnet}, pixel-wise planar prediction in GG-CNN~\cite{morrison2018ggcnn,morrison2020reactive}, descriptor-based category-level pose transfer in Neural Descriptor Fields (NDF)~\cite{simeonov2021ndf}, and six-degree-of-freedom (6-DoF) grasp generation from point clouds~\cite{mousavian2019graspnet,sundermeyer2021contact,wang2021graspness}. Large benchmarks and simulated grasp corpora support this progress~\cite{fang2020graspnet,eppner2021acronym}.

Geometric proposal methods detect parallel-jaw grasps directly in point clouds~\cite{gualtieri2016highprecision,tenpas2017grasp}, while AnyGrasp extends dense grasp perception across spatial and temporal domains~\cite{fang2023anygrasp}. GraspSAM integrates promptable segmentation with grasp prediction~\cite{noh2024graspsam}. RAGrasp instead stores successful grasps explicitly and focuses on retrieval, transfer, and geometric validation. Its compact memory is collected on the target setup, mitigating the sensor, embodiment, and scene mismatch between public training corpora and local deployment. This trades universal open-world coverage for deployment-specific precision and data-efficient adaptation. The approaches remain complementary: a learned detector may generate initial templates, while retrieval supplies inspectable instance-level memory.

\subsection{Foundation Features and Promptable Segmentation}
DINO (self-distillation with no labels) showed that a self-supervised Vision Transformer can organize image patches according to object and part structure without task-specific annotations~\cite{caron2021dino}. DINOv2 extends this approach and provides two types of descriptors used in this work: a global image descriptor that summarizes the masked object and patch descriptors that retain local spatial information~\cite{oquab2024dinov2}. Objects with similar appearance or parts tend to have similar descriptors, making the representation useful for retrieval and correspondence.

The Segment Anything Model (SAM) is a promptable segmentation network: given an image and a small set of positive points or a bounding box, it returns a pixel-level object mask~\cite{kirillov2023sam}. SAM2 extends this formulation while retaining the same prompt-to-mask interface~\cite{ravi2025sam2}. RAGrasp uses DINOv2 features to help locate the object and to compare the query with stored templates; SAM2 converts the fused object proposal into the mask used by all later geometric operations. For local matching, classical methods use the scale-invariant feature transform (SIFT)~\cite{lowe2004sift}, whereas SuperGlue and LightGlue learn to match detected image features~\cite{sarlin2020superglue,lindenberger2023lightglue}. RAGrasp instead compares DINOv2 patch descriptors and retains pairs that are mutual nearest neighbors.

\subsection{Geometric Registration and Grasp Transfer}
Iterative closest point (ICP) aligns two geometric point sets by alternating between nearest-neighbor assignment and least-squares transform estimation~\cite{besl1992icp}; efficient variants improve correspondence selection and convergence behavior~\cite{rusinkiewicz2001icp}. Local 3D descriptors such as fast point feature histograms (FPFH) can further support geometric matching~\cite{rusu2009fpfh}. Random sample consensus (RANSAC) estimates a model from repeatedly sampled correspondence subsets and rejects inconsistent matches as outliers~\cite{fischler1981ransac}. Given an accepted correspondence set, Umeyama's closed-form estimator recovers the least-squares rotation, translation, and uniform scale~\cite{umeyama1991transformation}. RAGrasp combines these tools rather than relying on a single estimator: contour ICP provides a conservative geometry-based transfer, while DINOv2 patch correspondences and RANSAC provide a semantics-aware alternative. Metric execution additionally relies on intrinsic camera calibration~\cite{zhang2000calibration} and camera-to-robot hand-eye calibration~\cite{tsai1989handeye}.

\section{Problem Formulation}
Let the query observation be an RGB image $I_q$, an aligned depth map $D_q$, camera intrinsics $\mathbf{K}$, and the camera-to-robot homogeneous transform ${}^{R}\mathbf{T}_{C}$, whose superscript and subscript denote the destination and source frames, respectively. A planar parallel-jaw grasp is
\begin{equation}
  \mathbf{g}=(u,v,\theta,w),
\end{equation}
where $(u,v)$ is the grasp center in image pixels, $\theta$ is the angle of the gripper closing direction measured from the image $u$-axis, and $w$ is the image-space jaw opening. More precisely, $w$ is the pixel distance between the two inner jaw-contact markers in the 2D grasp representation; it is converted to a metric opening only after depth back-projection in Stage~5. The template bank is
\begin{equation}
  \mathcal{B}=\{(I_i,M_i,\mathbf{g}_i,\mathbf{f}_i)\}_{i=1}^{N},
\end{equation}
where $I_i$ is the template RGB image, $M_i$ is its binary object mask, $\mathbf{g}_i$ is its annotated grasp, and $\mathbf{f}_i$ contains cached contour samples and DINOv2 descriptors. The objective is to retrieve the most compatible template $i^*$, estimate a 2D transfer $\mathcal{T}_{i^*\rightarrow q}$ from that template to the query, and refine the transferred grasp to obtain $\mathbf{g}_q^*$. The result must remain supported by the query mask, provide approximately opposing silhouette contacts, and be convertible to a metric robot-relative pose.

The bank $\mathcal{B}$ is constructed from a compact set of locally captured RGB-D samples using the deployed sensing and manipulation stack. Each sample requires a grasp annotation and an object mask but does not require training a task-specific grasp network. Adaptation to a new site or object family is therefore performed by adding templates rather than by retraining the complete perception policy.

\section{Method}
Figure~\ref{fig:main-pipeline} defines the five processing stages used throughout this paper.

\subsection{Stage 1: Quality-Gated Semantic Object Mask}
DINOv2 divides the query image into patches and maps each patch to a feature vector, producing the dense feature map $\mathbf{F}_q$. In parallel, color contrast and the residual from an estimated tabletop depth plane identify pixels that differ from the local background in appearance or height. The three cues are fused into positive points or a bounding box and supplied to SAM2, which returns one or more candidate object masks. For a candidate mask $M_q$, the quality gate checks that the mask contains the fused positive evidence, has sufficient foreground area, and follows a depth discontinuity rather than a background texture boundary. A rejected mask terminates the current query; only an accepted mask is passed to retrieval. This ordering prevents background regions from contaminating contour matching and masked DINOv2 descriptors.

\subsection{Stage 2: Geometry-Semantic Template Retrieval}
Retrieval uses a coarse-to-fine cascade for computational efficiency. Contour alignment is applied to all $N$ templates because it is substantially cheaper than dense DINOv2 patch comparison. The more expensive semantic scores are then evaluated only for the $K$ most promising geometric candidates. For template $i$, scaled multi-start contour ICP estimates a 2D similarity transform $\mathbf{A}^{\mathrm{ICP}}_i$---rotation, translation, and uniform scale---between the boundaries of $M_i$ and $M_q$. Its coarse score is
\begin{equation}
 s_i^{\mathrm{geo}} = \alpha\,\mathrm{IoU}(\mathbf{A}^{\mathrm{ICP}}_i M_i,M_q)
 - \beta\,\mathrm{RMSE}_{\mathrm{ctr}}(i),
 \label{eq:geometry-score}
\end{equation}
where $\mathrm{IoU}$ is the intersection-over-union of the aligned masks, $\mathrm{RMSE}_{\mathrm{ctr}}(i)$ is the root-mean-square distance between corresponding contour points after alignment, and $\alpha,\beta>0$ balance overlap against boundary error. All templates are ranked by $s_i^{\mathrm{geo}}$. The shortlist is the index set
\begin{equation}
 \mathcal{S}_K=\operatorname{TopK}_{i\in\{1,\ldots,N\}}s_i^{\mathrm{geo}},
 \label{eq:shortlist}
\end{equation}
where $K$ is a fixed retrieval hyperparameter. Thus, ``shortlisted'' means that a template is among the $K$ largest coarse-score values; no additional score threshold is used.

Only templates in $\mathcal{S}_K$ undergo masked DINOv2 comparison. This second stage calculates three complementary similarities: $s_i^{\mathrm{global}}$ is the cosine similarity between the mean masked DINOv2 descriptors; $s_i^{\mathrm{dense}}$ is the average patchwise cosine similarity after warping the template with $\mathbf{A}^{\mathrm{ICP}}_i$; and $s_i^{\mathrm{match}}$ measures the fraction and consistency of mutual-nearest-neighbor patch correspondences. Their weighted combination is the fine score
\begin{equation}
 s_i^{\mathrm{fine}} = \lambda_g s_i^{\mathrm{global}}
 +\lambda_d s_i^{\mathrm{dense}}
 +\lambda_m s_i^{\mathrm{match}},
 \label{eq:fine-score}
\end{equation}
where $\lambda_g,\lambda_d,\lambda_m\geq0$ control the contribution of the three terms. The retrieved template is selected only by the fine score within the shortlist,
\begin{equation}
 i^*=\arg\max_{i\in\mathcal{S}_K}s_i^{\mathrm{fine}}.
 \label{eq:template-selection}
\end{equation}
The coarse score therefore reduces the number of expensive semantic comparisons, whereas the fine score makes the final retrieval decision.

For the analysis in Fig.~\ref{fig:similarity}, we define the object-template similarity of a query as the largest aligned-mask overlap available in the bank,
\begin{equation}
 \rho_q=\max_{i\in\{1,\ldots,N\}}
 \mathrm{IoU}(\mathbf{A}^{\mathrm{ICP}}_iM_i,M_q).
 \label{eq:object-similarity}
\end{equation}
This diagnostic quantity measures geometric coverage of the query by the template bank; it is not a new semantic class label.

\subsection{Stage 3: Confidence-Gated Transfer}
For the retrieved template $i^*$, RAGrasp computes two candidate transfers but applies exactly one of them. The geometry path refines the mask scale and contour ICP alignment, producing $\mathbf{A}_{g}$. In parallel, the semantic path forms mutual-nearest-neighbor DINOv2 patch correspondences, rejects inconsistent pairs with RANSAC, and estimates a 2D similarity transform $\mathbf{A}_{s}$ from the remaining inliers.

The semantic confidence combines three normalized quantities. Let $r_{\mathrm{in}}\in[0,1]$ be the fraction of tentative correspondences accepted as RANSAC inliers, $\bar e_s\in[0,1]$ the normalized mean reprojection error of those inliers, and $a_s\in[0,1]$ their spatial coverage within the query mask. We calculate
\begin{equation}
 c_s=\gamma_r r_{\mathrm{in}}+\gamma_e(1-\bar e_s)+\gamma_a a_s,
 \label{eq:semantic-confidence}
\end{equation}
where $\gamma_r,\gamma_e,\gamma_a\geq0$ and $\gamma_r+\gamma_e+\gamma_a=1$. A high $c_s$ therefore requires numerous, geometrically consistent, and spatially distributed semantic matches. The gate compares $c_s$ with the fixed threshold $\tau_c$ and selects one transfer:
\begin{equation}
 \mathbf{A}^{*}=\begin{cases}
 \mathbf{A}_{s}, & c_s\geq \tau_c,\\
 \mathbf{A}_{g}, & c_s<\tau_c.
 \end{cases}
 \label{eq:transfer-gate}
\end{equation}
Here, $\mathbf{A}_{s}$ is the semantic-correspondence transform, $\mathbf{A}_{g}$ is the contour-based transform, $c_s$ is semantic confidence, $\tau_c$ is the decision threshold, and $\mathbf{A}^{*}$ is the single transform passed to the next stage. The two transforms are never averaged: the semantic path is used when its evidence passes the gate, and the geometry path is the fallback otherwise. The template grasp center and two jaw markers are transformed by $\mathbf{A}^{*}$, from which the coarse center, angle, and opening are recomputed. Transforming the jaw markers, rather than only the center, preserves the scale and orientation induced by the selected path.

\subsection{Stage 4: Mask-Constrained Local Refinement}
The coarse grasp is first snapped to the foreground mask if its center falls outside the accepted object region. Local mask principal components provide a stable orientation cue, while the aligned template supplies the transferred prior. Around the coarse grasp, the system searches perturbations $\Delta\mathbf{g}=(\Delta u,\Delta v,\Delta\theta,\Delta w)$ and evaluates mask support and antipodal contact quality:
\begin{align}
 \mathbf{g}_q^*&=\arg\max_{\mathbf{g}\in\mathcal{N}(\mathbf{g}_q^0)}J(\mathbf{g}),\\
 J(\mathbf{g})&=\eta_m S_{\mathrm{mask}}(\mathbf{g})+
 \eta_a S_{\mathrm{anti}}(\mathbf{g})
 -\eta_p\lVert\mathbf{g}-\mathbf{g}_q^0\rVert_{\Sigma}^{2}.
 \label{eq:local-refinement}
\end{align}
Here, $\mathcal{N}(\mathbf{g}_q^0)$ is a bounded neighborhood around the transferred grasp $\mathbf{g}_q^0$. The mask term $S_{\mathrm{mask}}$ rewards a grasp center and finger approach region supported by $M_q$; $S_{\mathrm{anti}}$ rewards two contact markers on opposite mask boundaries with approximately opposing boundary normals. The final term penalizes deviation from the transferred grasp, with $\boldsymbol{\Sigma}$ normalizing the different units of pixel displacement, angle, and jaw opening. The non-negative weights $\eta_m$, $\eta_a$, and $\eta_p$ control these three objectives. The search jointly adjusts center, angle, and opening. After a verified successful execution, the refined query mask and grasp can be written back to the template bank, expanding coverage without full-network retraining.

\subsection{Stage 5: Metric Robot Command and Execution}
The refined grasp is still expressed in image coordinates: $(u^*,v^*)$ and $w^*$ are measured in pixels, not meters. The center is back-projected using the aligned depth $z=D_q(u^*,v^*)$:
\begin{equation}
 {}^{C}\mathbf{p}=z\mathbf{K}^{-1}[u^*,v^*,1]^{\mathsf T},\qquad
 \begin{bmatrix}{}^{R}\mathbf{p}\\1\end{bmatrix}
 ={}^{R}\mathbf{T}_{C}
 \begin{bmatrix}{}^{C}\mathbf{p}\\1\end{bmatrix}.
 \label{eq:backprojection}
\end{equation}
where $\mathbf{K}$ contains the camera intrinsics, ${}^{C}\mathbf{p}$ is the 3D grasp center expressed in the camera frame, ${}^{R}\mathbf{T}_{C}$ maps a point from the camera frame $C$ to the robot base frame $R$, and ${}^{R}\mathbf{p}$ is the same center expressed in the robot base frame. The two jaw markers are back-projected and transformed using the same notation; their 3D separation gives the metric gripper opening corresponding to the pixel distance $w^*$. The in-plane angle $\theta^*$ is combined with the calibrated camera-to-robot orientation to obtain the end-effector yaw. The resulting robot-relative grasp pose is sent to the controller. An execution is accepted only if the object is lifted by $20\,\mathrm{cm}$ from its pre-grasp height and remains held for $10\,\mathrm{s}$ without being dropped. Only an accepted execution is eligible for template write-back.

\begin{figure}[t]
  \centering
  \includegraphics[width=\linewidth]{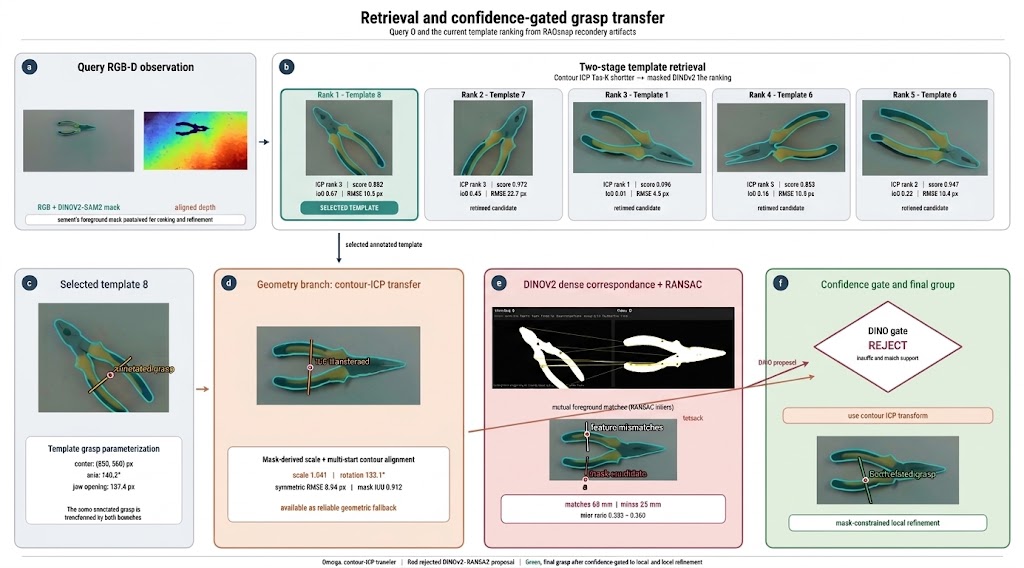}
  \caption{Qualitative example of geometry-semantic retrieval and confidence-gated dual-path grasp transfer.}
  \label{fig:transfer}
\end{figure}

\begin{figure*}[t]
  \centering
  \includegraphics[width=\textwidth]{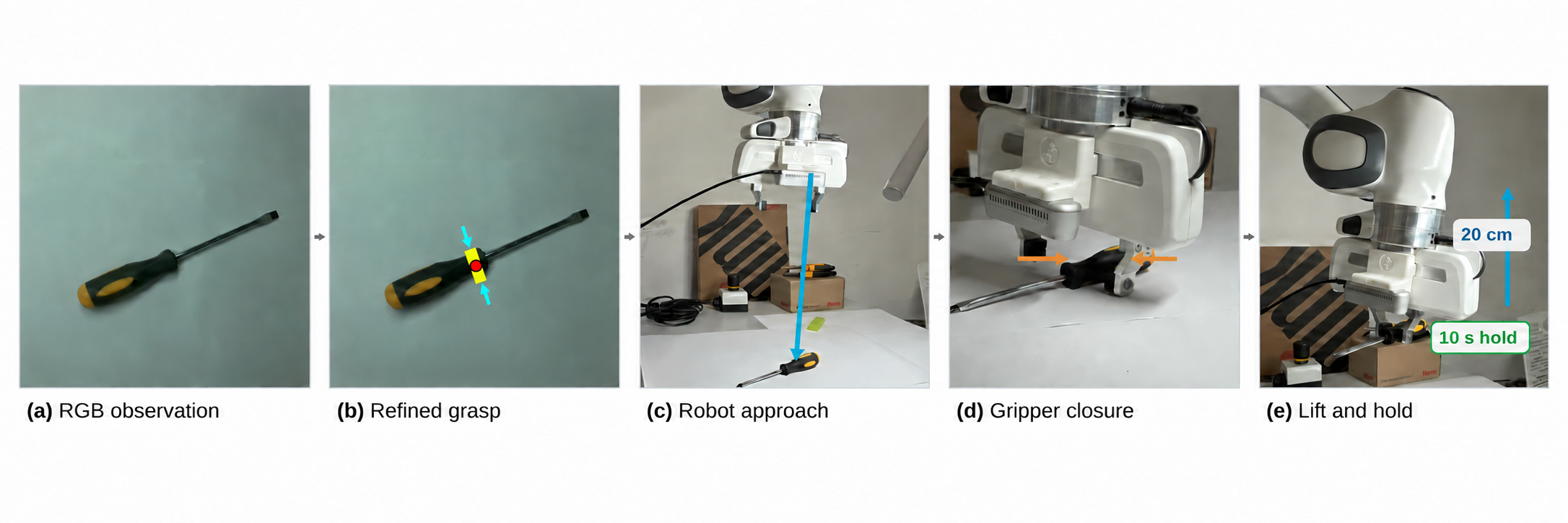}
  \caption{End-to-end grasp prediction and real-robot execution. (a) RGB observation of the target object. (b) Refined image-space grasp: the center marker represents $(u,v)$, the orientation indicator shows the closing direction $\theta$, and the separation between the two jaw markers represents the opening $w$. The gripper closes along the line joining the two jaw markers. (c) Robot approach to the calibrated target pose. (d) Parallel-gripper closure. (e) Successful execution verified by a 20-cm lift and a 10-s hold.}
  \label{fig:execution}
\end{figure*}

\section{Experiments}
\subsection{Evaluation Protocol}
\subsubsection{Hardware and data}
The physical setup comprises a Franka Emika Panda manipulator, its parallel-jaw gripper, and an Intel RealSense D435i RGB-D camera observing the tabletop workspace. The camera provides spatially aligned color and depth images to the perception pipeline. All template and query observations are captured with this same setup, so the stored examples contain the sensor noise, gripper geometry, and camera viewpoint encountered at deployment. No public grasp dataset is used for task-specific training; DINOv2 and SAM2 are used with publicly available pretrained weights. A \emph{seen} test object has at least one observation in the template bank, although its test pose may differ from the stored pose. An \emph{unseen} test object has no observation in the bank.

\subsubsection{Trial procedure and success criterion}
Each trial follows the same sequence. The target object is placed on the tabletop, one aligned RGB-D observation is acquired, and RAGrasp returns one refined planar grasp. After 2D-to-3D conversion, the robot approaches the grasp center from above, closes the parallel jaws to the commanded metric width, and lifts vertically. A trial is successful only when the gripper raises the object by $20\,\mathrm{cm}$ relative to its pre-grasp height and then holds it for $10\,\mathrm{s}$ without a drop. A failed closure or a drop during the lift or hold is counted as a failure. Success rate (SR) is the number of successful trials divided by the number of trials in the corresponding condition.

\subsubsection{Evaluation conditions}
\begin{figure}[!t]
  \centering
  \includegraphics[width=\columnwidth]{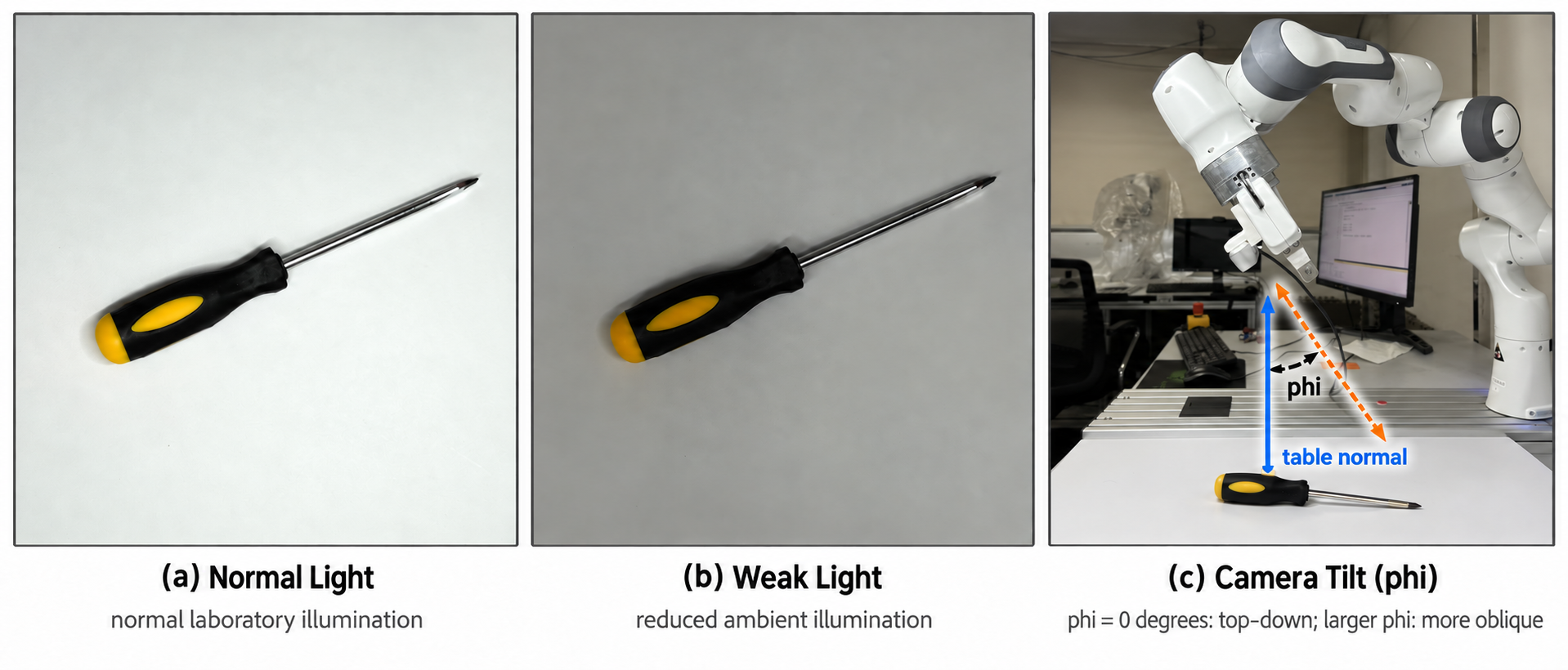}
  \caption{Conditions used in Table~\ref{tab:view-light}: (a) normal laboratory illumination; (b) reduced ambient illumination; and (c) camera tilt. The angle $\phi$ is measured between the viewing direction (orange) and the tabletop normal (blue), with $\phi=0^\circ$ denoting a top-down view. Arrows are schematic.}
  \label{fig:table1-conditions}
\end{figure}

We evaluate five aspects: (i) robustness to camera viewpoint and illumination on unseen objects; (ii) comparison with GraspSAM, NDF, and AnyGrasp on seen and unseen objects; (iii) sensitivity to object-template similarity; (iv) sensitivity to template-bank size; and (v) component ablations. Unless stated otherwise, experiments use normal laboratory illumination and a perpendicular camera view. ``Weak light'' denotes the same workspace under reduced ambient illumination. For the viewpoint study, the angle $\phi$ is the deviation of the camera viewing direction from the tabletop normal: $\phi=0^\circ$ is a perpendicular, top-down view, and larger values are progressively more oblique views. Figure~\ref{fig:table1-conditions} illustrates the illumination conditions and the geometric definition of $\phi$ used in Table~\ref{tab:view-light}. Each viewpoint-lighting condition contains 10 trials, one for each of 10 unseen objects. The baseline comparison uses 20 seen and 20 unseen objects under the same success criterion. All methods receive observations from the same sensor setup and execute one selected grasp with the same robot controller.

\subsection{Viewpoint and Illumination Robustness}
Table~\ref{tab:view-light} summarizes the viewpoint study under the conditions illustrated in Fig.~\ref{fig:table1-conditions}. Performance remains high across the tested camera tilts under normal illumination, whereas weak illumination causes additional failures at oblique views. This trend is consistent with reduced appearance contrast and less reliable semantic correspondence under darker imaging conditions.

\begin{table}[!t]
  \centering
  \caption{Physical grasp success under viewpoint and illumination changes. Each condition contains 10 trials.}
  \label{tab:view-light}
  \footnotesize
  \setlength{\tabcolsep}{4.2pt}
  \renewcommand{\arraystretch}{1.10}
  \begin{tabular}{@{}ccc@{}}
    \toprule
    \textbf{Camera Tilt $\phi$} & \textbf{Normal Light} & \textbf{Weak Light} \\
    \midrule
    $0^\circ$  & 10/10 (100\%) & 10/10 (100\%) \\
    $15^\circ$ & 9/10 (90\%) & 8/10 (80\%) \\
    $30^\circ$ & 9/10 (90\%)   & 9/10 (90\%) \\
    $45^\circ$ & 9/10 (90\%)   & 8/10 (80\%) \\
    \bottomrule
  \end{tabular}
\end{table}
\FloatBarrier

\subsection{Baseline Comparison}
Table~\ref{tab:baselines} compares RAGrasp with three representative learning- and transfer-based approaches under the common protocol described above. GraspSAM extends the Segment Anything Model with lightweight adapters, learnable grasp tokens, and a modified decoder to jointly support prompt-driven object segmentation and category-agnostic grasp prediction~\cite{noh2024graspsam}. Neural Descriptor Fields (NDF) learn self-supervised, $\mathrm{SE}(3)$-equivariant descriptors from 3D observations and transfer a demonstrated gripper pose to a new object instance by optimizing for a matching pose descriptor~\cite{simeonov2021ndf}. AnyGrasp is a point-cloud-based grasp-perception method that generates dense 7-DoF grasp poses and incorporates spatial-temporal supervision, center-of-mass awareness, and grasp correspondence for robust static and dynamic grasping~\cite{fang2023anygrasp}. These baselines respectively represent promptable image-based grasp detection, descriptor-based 3D pose transfer, and dense 3D grasp generation, whereas RAGrasp retrieves a locally collected RGB-D template and explicitly transfers its annotated grasp.

All four methods are competitive on seen objects. Under this experimental protocol, RAGrasp attains the highest observed success rate on unseen objects, indicating that local template retrieval remains effective when the exact test instance is absent from memory.

\begin{table}[!t]
  \centering
  \caption{Physical grasp success under normal lighting and a perpendicular camera view. Each split contains 20 objects.}
  \label{tab:baselines}
  \footnotesize
  \setlength{\tabcolsep}{4.2pt}
  \renewcommand{\arraystretch}{1.10}
  \begin{tabular}{@{}lcc@{}}
    \toprule
    \textbf{Method} & \textbf{Seen Objects} & \textbf{Unseen Objects} \\
    \midrule
    \textbf{RAGrasp} & \textbf{20/20 (100\%)} & \textbf{19/20 (95\%)} \\
    GraspSAM         & 18/20 (90\%)          & 16/20 (80\%) \\
    NDF       & 18/20 (90\%)          & 17/20 (85\%) \\
    AnyGrasp         & 20/20 (100\%)         & 16/20 (80\%) \\
    \bottomrule
  \end{tabular}
\end{table}
\FloatBarrier

\subsection{Template Coverage}
\begin{figure}[!t]
  \centering
  \includegraphics[width=\columnwidth]{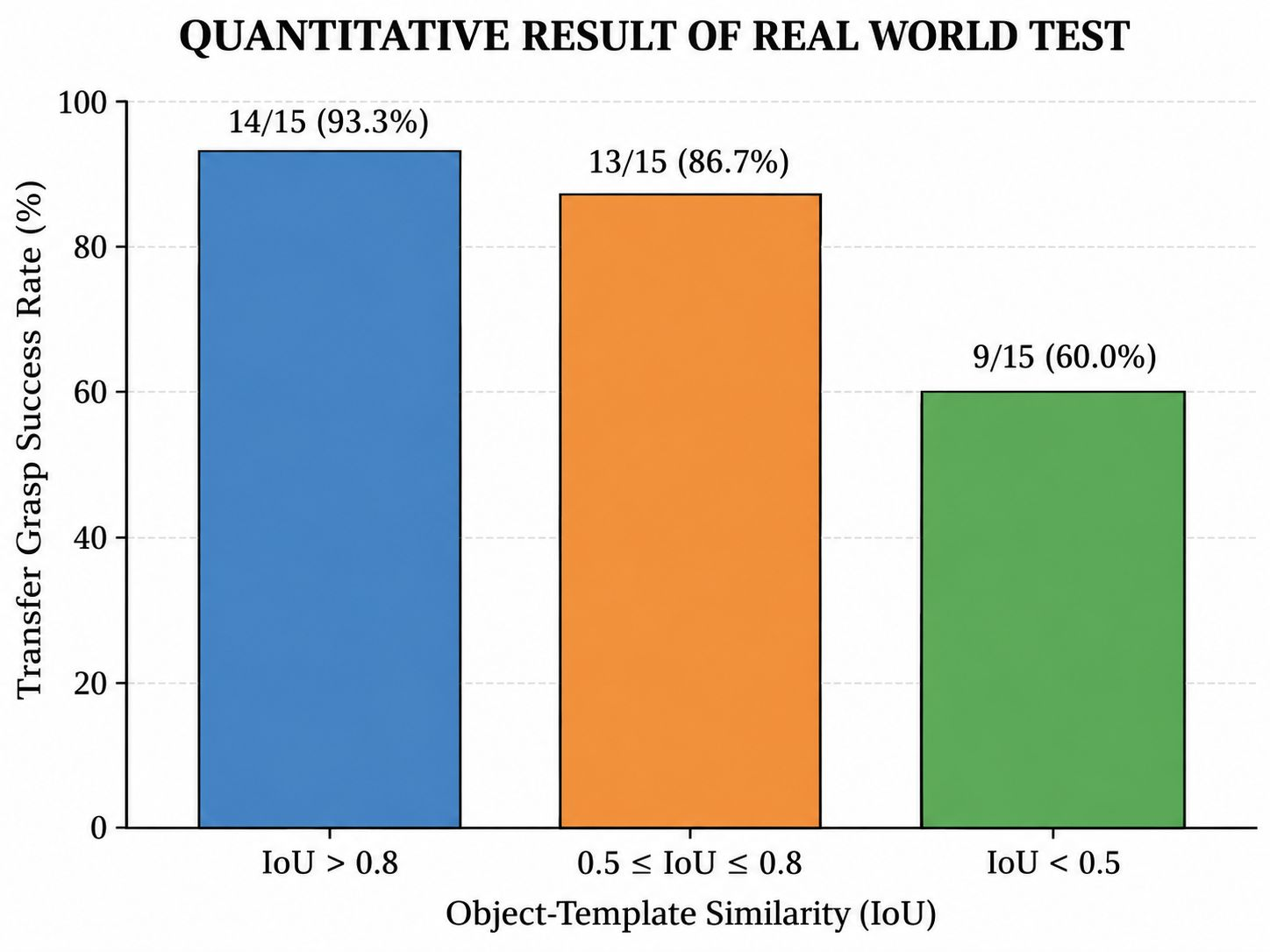}
  \caption{Transfer-grasp success versus object-template similarity $\rho_q$, defined in~\eqref{eq:object-similarity} as the maximum contour-aligned mask IoU between the query and any template. Performance decreases as geometric coverage of the query by the bank becomes weaker.}
  \label{fig:similarity}
\end{figure}

\begin{figure}[!t]
  \centering
  \includegraphics[width=\columnwidth]{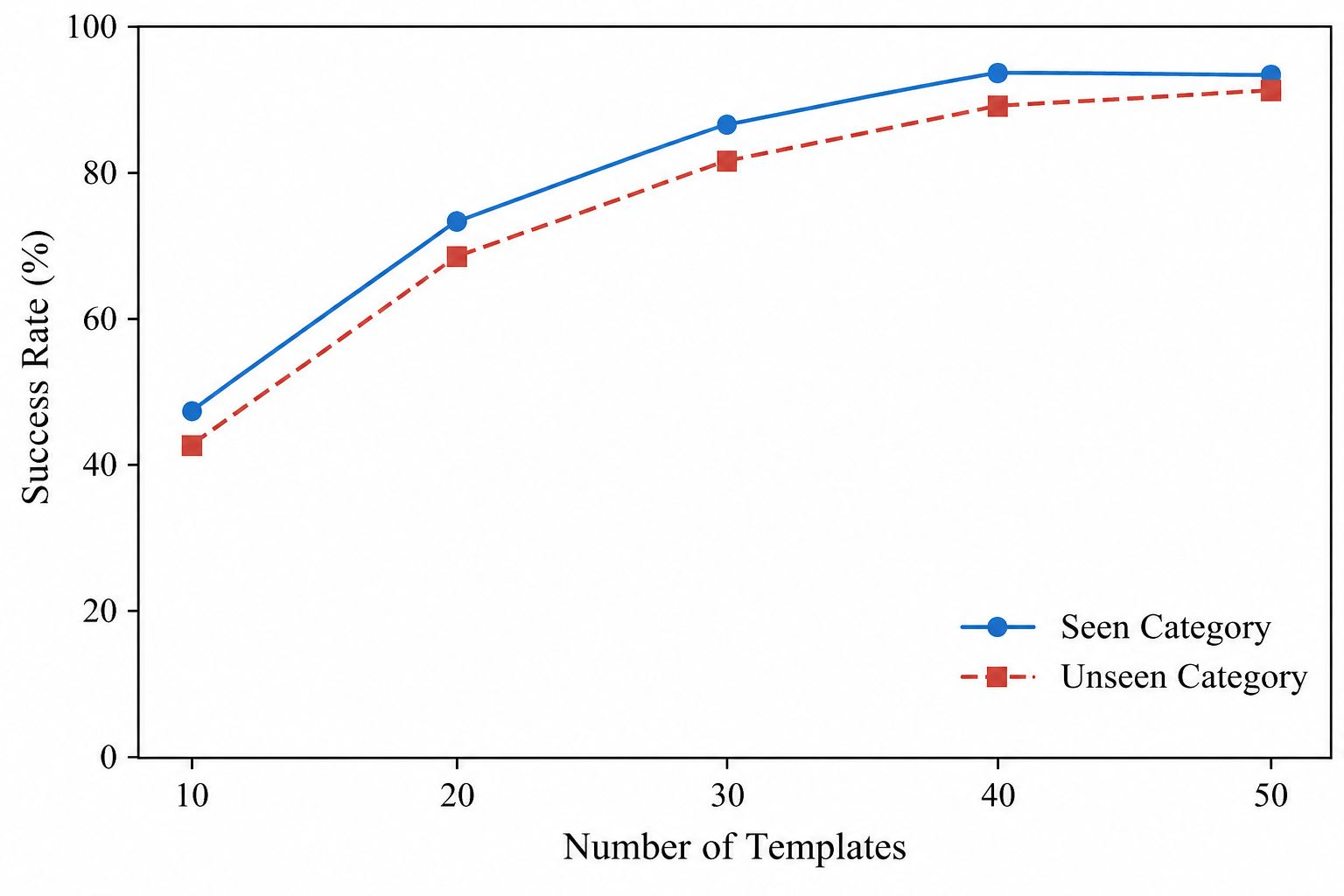}
  \caption{Success rate versus the number of templates in the retrieval bank. Seen and unseen performance improves rapidly from 10 to 40 templates and then approaches saturation.}
  \label{fig:bank-size}
\end{figure}

Figure~\ref{fig:similarity} groups trials by the object-template similarity $\rho_q$ introduced in~\eqref{eq:object-similarity}. SR is 14/15 (93.3\%) when $\rho_q>0.8$, 13/15 (86.7\%) for $0.5\leq\rho_q\leq0.8$, and 9/15 (60.0\%) for $\rho_q<0.5$. Thus, $\rho_q$ measures whether the bank contains a shape that can support reliable transfer; it is distinct from the DINOv2 semantic similarity used for fine ranking. Figure~\ref{fig:bank-size} provides a complementary bank-size study. Performance increases sharply between 10 and 30 templates for both seen and unseen categories, improves further at 40 templates, and changes only marginally at 50. Together, these results indicate that memory diversity contributes most until the bank covers the dominant query geometries.

\subsection{Ablation Study}
Table~\ref{tab:ablation} reports the effect of removing individual components from the pipeline in Fig.~\ref{fig:main-pipeline}. Geometry-only transfer achieves 80\% SR on both splits. Always selecting DINOv2-RANSAC without confidence-gated fallback yields 70\% unseen SR. Replacing the DINOv2+SAM2 semantic mask with an RGB-D geometric mask produces the lowest unseen SR, 65\%. Removing mask-constrained local refinement yields 80\% unseen SR. Disabling success-conditioned write-back reduces seen and unseen SR to 90\% and 80\%, respectively, indicating that retaining verified executions expands useful template coverage. The complete pipeline achieves 100\% SR on seen objects and 95\% on unseen objects under this protocol.

\begin{table}[!t]
  \centering
  \caption{Ablation study on seen and unseen categories. Each entry reports successful grasps over 20 trials.}
  \label{tab:ablation}
  \scriptsize
  \setlength{\tabcolsep}{2.5pt}
  \renewcommand{\arraystretch}{1.08}
  \begin{tabular}{@{}>{\raggedright\arraybackslash}p{0.53\columnwidth}cc@{}}
    \toprule
    \textbf{Ablated Version} & \textbf{Seen} & \textbf{Unseen} \\
    \midrule
    Geometry-only transfer\newline{(w/o DINOv2 correspondence and confidence gate)}
      & 16/20 (80\%) & 16/20 (80\%) \\
    \addlinespace
    w/o confidence-gated fallback\newline{(always use DINOv2-RANSAC transfer)}
      & 17/20 (85\%) & 14/20 (70\%) \\
    \addlinespace
    w/o DINOv2+SAM2 semantic mask\newline{(use RGB-D geometric mask)}
      & 15/20 (75\%) & 13/20 (65\%) \\
    \addlinespace
    w/o mask-constrained local refinement\newline{(use transferred grasp directly)}
      & 18/20 (90\%) & 16/20 (80\%) \\
    \addlinespace
    w/o success-conditioned write-back\newline{(do not add verified executions to the template bank)}
      & 18/20 (90\%) & 16/20 (80\%) \\
    \midrule
    \textbf{Full RAGrasp} & \textbf{20/20 (100\%)} & \textbf{19/20 (95\%)} \\
    \bottomrule
  \end{tabular}
\end{table}

\section{Discussion and Limitations}
RAGrasp is most effective when the template bank contains a geometrically related instance and when either contour alignment or semantic correspondence remains reliable. The confidence gate mitigates failure of a single transfer estimator, but it cannot compensate for missing shape coverage. The 60.0\% SR in the lowest-similarity group ($\rho_q<0.5$) reflects this limitation. Performance also depends on mask quality, aligned depth, and camera-to-robot calibration. The evaluation uses 20 objects per split and does not report confidence intervals or end-to-end latency; the conclusions should therefore be interpreted within this scope. Additional repetitions, scene diversity, and runtime characterization would provide a more complete assessment.

\section{Conclusion}
This paper introduced RAGrasp, a five-stage geometry-semantic pipeline for retrieving, transferring, refining, and executing planar grasps from a compact, locally collected template bank. By adapting stored local experience instead of training a task-specific predictor on a large public grasp corpus, the method avoids task-specific deployment retraining and requires limited local annotation. Contour ICP and DINOv2 provide complementary retrieval and transfer evidence; a confidence gate selects between them; SAM2 masks and silhouette-based contact search constrain local refinement; and calibrated depth converts the result into a robot-relative pose. In the evaluated setting, RAGrasp achieves high seen- and unseen-object success rates while remaining dependent on template coverage.

\begingroup
\ifieeetran\else\small\fi

\endgroup

\end{document}